\documentclass{article}

\usepackage{microtype}
\usepackage{graphicx}
\usepackage{subfigure}
\usepackage{booktabs} % for professional tables

\usepackage{hyperref}

\usepackage[accepted]{icml2023}
\usepackage{amsmath}
\usepackage{amssymb}
\usepackage{mathtools}
\usepackage{amsthm}

\usepackage[capitalize,noabbrev]{cleveref}

\theoremstyle{plain}

\theoremstyle{definition}

\theoremstyle{remark}

\usepackage[textsize=tiny]{todonotes}

\icmltitlerunning{An Empirical Study of the Effectiveness of Using a Replay Buffer on Mode Discovery in GFlowNets}

\begin{document}

\twocolumn[
\icmltitle{An Empirical Study of the Effectiveness of Using a Replay Buffer on Mode Discovery in GFlowNets}



\icmlsetsymbol{equal}{*}

\begin{icmlauthorlist}
\icmlauthor{Nikhil Vemgal}{mila,mcgill}
\icmlauthor{Elaine Lau}{mila,mcgill}
\icmlauthor{Doina Precup}{mila,mcgill,deepmind}
% \icmlauthor{Firstname1 Lastname1}{equal,yyy}
% \icmlauthor{Firstname2 Lastname2}{equal,yyy,comp}
% \icmlauthor{Firstname3 Lastname3}{comp}
\end{icmlauthorlist}

\icmlaffiliation{mila}{Mila, Montreal, Canada}
\icmlaffiliation{mcgill}{School of Computer Science, McGill University, Montreal, Canada}
\icmlaffiliation{deepmind}{DeepMind, Montreal, Canada}

\icmlcorrespondingauthor{Nikhil Vemgal}{nikhil.vemgal@mail.mcgill.ca}

\icmlkeywords{Machine Learning, ICML}

\vskip 0.3in
]



\printAffiliationsAndNotice{}  % leave blank if no need to mention equal contribution

\begin{abstract}

Reinforcement Learning (RL) algorithms aim to learn an optimal policy by iteratively sampling actions to learn how to maximize the total expected return, $R(x)$. GFlowNets are a special class of algorithms designed to generate diverse candidates, $x$, from a discrete set, by learning a policy that approximates the proportional sampling of $R(x)$. GFlowNets exhibit improved mode discovery compared to conventional RL algorithms, which is very useful for applications such as drug discovery and combinatorial search. However, since GFlowNets are a relatively recent class of algorithms, many techniques which are useful in RL have not yet been associated with them. In this paper, we study the utilization of a replay buffer for GFlowNets. We explore empirically various replay buffer sampling techniques and assess the impact on the speed of mode discovery and the quality of the modes discovered. Our experimental results in the Hypergrid toy domain and a molecule synthesis environment demonstrate significant improvements in mode discovery when training with a replay buffer, compared to training only with trajectories generated on-policy.

\end{abstract}

\section{Introduction}
 Generative Flow Networks (GFlowNets) \cite{bengio2021flow} are a class of reinforcement learning (RL) algorithms which have been recently proposed, whose goal is to learn a stochastic policy to generate diverse objects from a discrete set, such as graphs. This is achieved by iteratively sampling actions that construct the object through a sequence of edits. GFlowNets \cite{bengio2021gflownet} sample a diverse set of objects $x$ with a training objective that approximately samples $x$ in proportion to the reward function $R(x)$ associated with it\footnote{Unlike in usual RL, rewards are only associated with objects $x$ corresponding to terminal states.}.

In drug discovery, the learner has access to an oracle, which is periodically queried with a batch of candidate molecules, to obtain rewards that estimate the efficacy of each candidate \cite{bengio2021flow}. The oracle usually takes the form of a neural network, trained as a proxy for estimating binding affinity to a target protein. Given the inherent uncertainty involved in drug trials and the approximate nature of the proxy reward (which is learned in a supervised manner from available data), it is important to have a diverse set of candidates. 

Existing work has demonstrated that GFlowNets outperform traditional techniques like Bayesian Optimization, and Markov Chain Monte Carlo (MCMC) in terms of both training efficiency and the diversity of the candidates discovered \cite{bengio2021flow}. GFlowNet is an offline off-policy learning algorithm \cite{bengio2021flow}, but the training of GFlowNets has predominantly focused on utilizing the data generated by the stochastic policy which is trained. That is, for every training step, a fixed set of trajectories is sampled from the current policy and used to train the GFlowNet. 

In this paper, we conduct an empirical analysis to evaluate the impact of adding experience replay to the GFlowNet training process. We examine three different training configurations, (i) without replay buffer, (ii) with a replay buffer that uses random sampling to choose training tuples, and (iii) R-PRS (Reward Prioritized Replay Sampling), a technique inspired by Prioritized Experience Replay (PER)~\cite{schaul2015prioritized}. We perform experiments on a Hypergrid toy domain and on a large-scale molecular synthesis environment. The empirical results demonstrate that using a replay buffer with GFlowNets significantly improves the training speed, the diversity of generated candidates, and the ability to discover different modes of the distribution. 

\section{Preliminaries}
Let $G=(\mathcal{S}, \mathcal{A})$ be a directed acyclic graph (DAG) \cite{bengio2021flow, bengio2021gflownet}, where $\mathcal{S}$ is the set of states (vertices) and $\mathcal{A}$ is the set of actions (edges). In GFlowNets,  the learner constructs $G$ using a series of actions (edges) starting from an initial state, $s_0 \in \mathcal{S}$ until a terminal (sink) node, $s_n \in \mathcal{S}$ is reached. 
A \textit{complete trajectory} \cite{malkin2022trajectory}, $\tau$, is a sequence of transitions from  $s_0$ to a terminal state: $\tau = (s_0 \rightarrow s_1 \rightarrow \dots \rightarrow s_n )$, where $(s_i \rightarrow s_{i+1}) \in \mathcal{A} \text{ }, \forall i$. A \textit{trajectory flow} $F:\mathcal{T} \mapsto \mathbb{R}^+$ is any non-negative function defined on the set of complete trajectories $\mathcal{T}$ to $\mathbb{R}^+$. The trajectory flow can be interpreted as the total amount of unnormalized probability flowing through a state. More formally, for any state $s$, the state flow is defined as $F(s) = \sum_{\tau \in \mathcal{T}: s \in \tau} F(\tau)$, and for any edge ($s \rightarrow s'$), the edge flow is defined as 
\begin{equation}
    F(s \rightarrow s') = \sum_{\tau \in \mathcal{T}:(s\rightarrow s' \in \tau)} F(\tau)
\end{equation}
The \textit{terminal flow} is defined as the flow associated with the final transition $(s_i \rightarrow s_n)$, $F(s_i \rightarrow s_n)$. The intention is to make the total flow at state $s_n$ approximately equal to the reward $R(s_n)$. The \textit{forward transition probability}, $P_F$ for each step of the trajectory is defined as: 
\begin{equation}
    P_F = \frac{F(s\rightarrow s')}{F(s)}
\end{equation}
and the probability of visiting a terminal state is:
\begin{equation}
    P_F(s) = \frac{\sum_{\tau \in \mathcal{T}: s \in \tau} F(\tau)}{Z}
\end{equation}
where $Z$ is the total flow, $Z=\sum_{\tau \in \mathcal{T}}F(\tau)$.

\textbf{Flow Matching Objective} \cite{bengio2021flow}: 
The \textit{flow matching criterion} states that the sum of inflow from all the parents of a node should be equal to the total outflow to all the children of that node:
\begin{equation} \label{FMLoss}
    \mathcal{L}_{FM}(s;\theta) = \left(\log \frac{\sum_{s' \in \text{Parent}(s)} F_{\theta} (s' \rightarrow s)}{\sum_{s'' \in \text{Child}(s)} F_{\theta} (s \rightarrow s'')} \right)^2.
    \end{equation}
\citep{bengio2021flow} showed that these constraints can be converted into a temporal-difference (TD)-like objective \cite{sutton1988learning} which is then optimized with respect to the parameters of a function approximator, like a neural network.
GFlowNets approximate the \textit{edge flow} $F_{\theta}: \mathcal{A} \rightarrow \mathbb{R}^+$ with learnable parameters $\theta$, such that the terminal flow is roughly equal to the reward function $R(x)$. 
Trajectories for training $\theta$ are sampled from an exploratory policy $\tilde{\pi}$ with full support, learned by minimizing the flow-matching objective (\ref{FMLoss}). 

\textbf{Trajectory Balance Objective} \cite{malkin2022trajectory}: The flow-matching objective can suffer from inefficient credit assignment. To overcome this, an alternative was proposed by \citeauthor{malkin2022trajectory}, which leads to faster convergence. The trajectory balance objective is defined as:
\begin{equation}
    \mathcal{L}_{TB}(\tau; \theta) = \left( \log \frac{Z_{\theta} \prod_{s\rightarrow s' \in \tau} P_{F_{\theta}}(s'|s)}{R(x)} \right)^2
\end{equation}

\section{Experience Replay}
Experience replay has emerged as a very useful RL technique which can improve learning efficiency and stability \cite{lin1992self}. 
The traditional approach involves storing past experiences encountered by the agent in a buffer and replaying them, by randomly sampling batches of experiences during the training process. The randomization allows the agent to explore diverse transitions, leading to better exploration and improved learning. Furthermore, if experience replay is done at the level of state-action transitions, rather than full trajectories, it breaks the temporal correlations between transitions, which can have a stabilizing effect when the RL agent is using non-linear function approximation. \citeauthor{mnih2015human} demonstrated the effectiveness of experience replay in Deep Q-Networks (DQNs), achieving state-of-the-art performance on a wide range of Atari 2600 games. 

\citeauthor{schaul2015prioritized} proposed a technique that enhances the replay buffer, by assigning priorities to the experiences stored therein. The idea is to prioritize and sample experiences based on the potential that they will induce learning. Prioritized Experience Replay (PER) assigns higher priority to experiences that have a larger TD-error magnitude, indicating that more would be learned from replaying this experience. This approach helps the agent learn from the most informative and challenging experiences.

\section{Experiments}

Our goal is to investigate the impact of different experience replay techniques on the training process of GFlowNets. Specifically, we compare three approaches: (i) training only with samples from the current online policy; (ii) training with an experience replay buffer that contains both samples from the current policy and from past policies, and where  random sampling is used to select batches; and (iii) R-PRS (Reward Prioritized Replay Sampling), a technique inspired by prioritized experience replay. In R-PRS, we store and sample trajectories with the highest reward in the replay buffer. During the sampling process, the learner prioritizes the buffer according to this reward (instead of the TD-error like in PER). The underlying hypothesis is that by prioritizing and learning from the most promising trajectories, the agent can effectively explore the state space and improve learning performance. This idea is very similar in spirit to the initial work on replay by \citet{lin1992self}.

\subsection{Hypergrid}

We first analyze the effect of using experience replay with GFlowNets in Hypergrid, a toy domain presented by \citeauthor{bengio2021flow}, which allows easy control of the number of interesting modes of a distribution and of the ease with which these modes can be discovered. The environment is an $n$-dimensional hypercube grid of side length $H$, where the states are the cells of the hypercube. The agent always starts at coordinate $x=(0,0,\dots)$, and the allowed actions $a_i$ increase the coordinate $i$, up to $H$, upon which the episode terminates. A \textit{stop} action can also terminate the episode. There are many sequences of actions that lead to the same goal state, making this MDP a DAG. 

We use the codebase and architecture developed by \cite{bengio2021flow} as a foundation. For the GFlowNet model, we use an MLP as the reward approximator, with two hidden layers, each with $256$ hidden units. We train all the models with the \textit{Flow Matching} objective. We set the learning rate to $0.001$ and use the Adam optimizer \cite{kingma2014adam}. All the experiments are run on 5 independent seeds and the mean and standard error are reported in the plots.

\begin{figure}[ht]
\vskip 0.2in
\begin{center}
% \centerline{\includegraphics[width=\columnwidth]{Plots/true_r.pdf}}
\centerline{\includegraphics[scale=0.4]{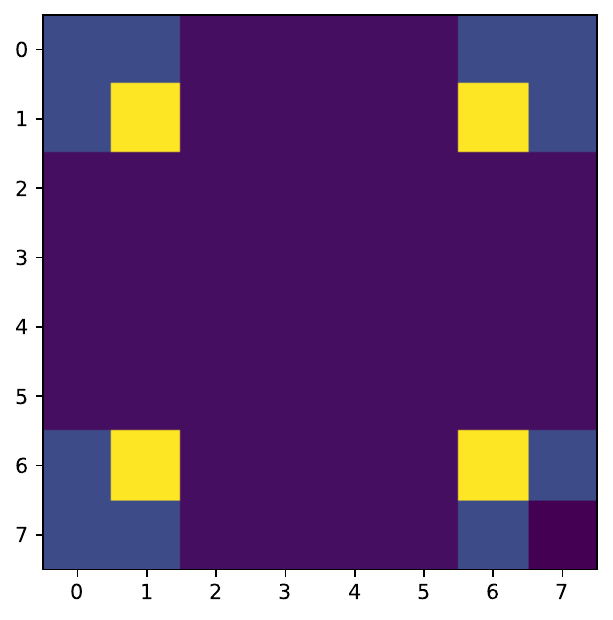}}
\caption{Hypergrid domain - Reward distribution for a 2-dimensional Hypergrid with $H=8$.}
\label{hypergrid_rewarddist}
\end{center}
\vskip -0.2in
\end{figure}

The reward for terminating the episode at coordinate $x$ is given by $R(x)>0$. We experimented with the reward function
$R(x) = R_0 + R_1 \prod_i \mathbb{I}(0.25 < |x_i/H - 0.5|) + R_2 \prod_i \mathbb{I}(0.3< |x_i / H -0.5| < 0.4)$ with $0<R_0 \ll R_1 < R_2$. We set $R_1=0.5$ and $R_2=2$. We vary the value of $R_0$ by setting it closer to 0, to make the problem artificially harder, by creating a region of state space which is less desirable to explore. The reward distribution for a 2D Hypergrid with $H=8$ is shown in Figure~\ref{hypergrid_rewarddist}. 

\begin{figure*}[h]
  \centering
  \begin{minipage}[t]{0.3\textwidth}
    \centering
    \includegraphics[width=\textwidth]{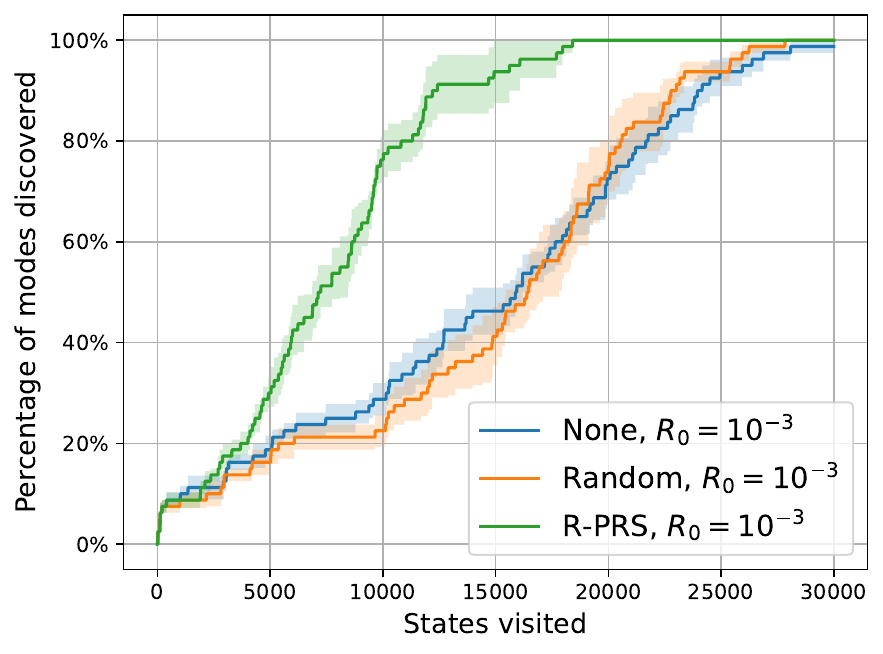}
    \vskip -0.1in
    \caption{Hypergrid domain - States visited vs. percentage of modes discovered during training in a 4-dimensional hypergrid (max = 16 modes) with $H=8$ for all three training regimes, with $R_0=10^{-3}$ (mean and standard error over 5 runs).}
    \label{hypergrid_1e-3_Allmethods}
    % \vspace{-50pt}
  \end{minipage}
  % \vskip -0.2in
  \hfill
  \begin{minipage}[t]{0.3\textwidth}
    \centering
    \includegraphics[width=\textwidth]{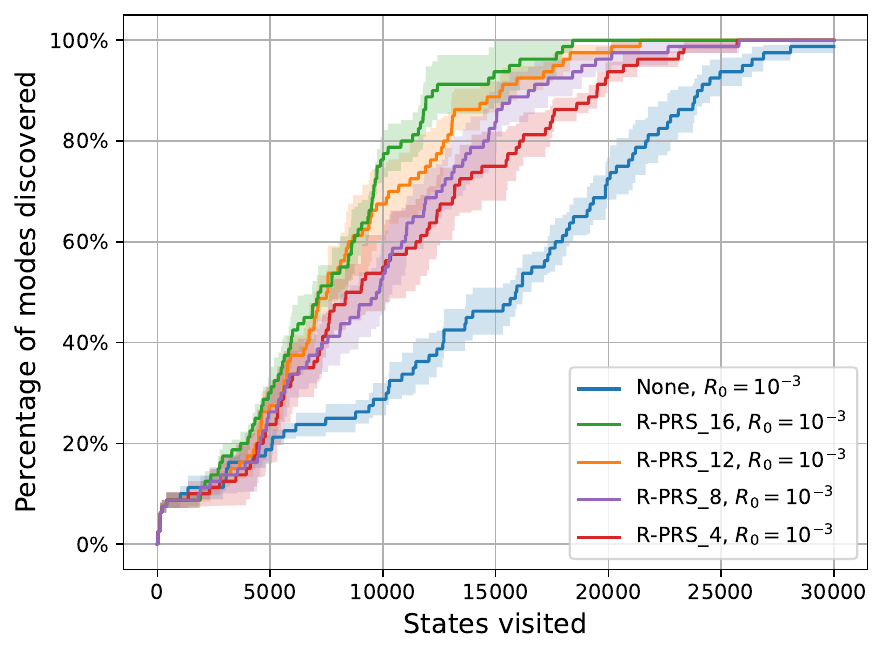}
    \vskip -0.1in
    \caption{Hypergrid domain - States visited vs. percentage of modes discovered during training in a 4-dimensional hypergrid (max = 16 modes) with $H=8$ for varying sample sizes for the R-PRS replay technique, with $R_0=10^{-3}$ (mean and standard error over 5 runs).}
    \label{hypergrid_1e-3_RPRS}
  \end{minipage}\hfill
  \begin{minipage}[t]{0.3\textwidth}
    \centering
    \includegraphics[width=\textwidth]{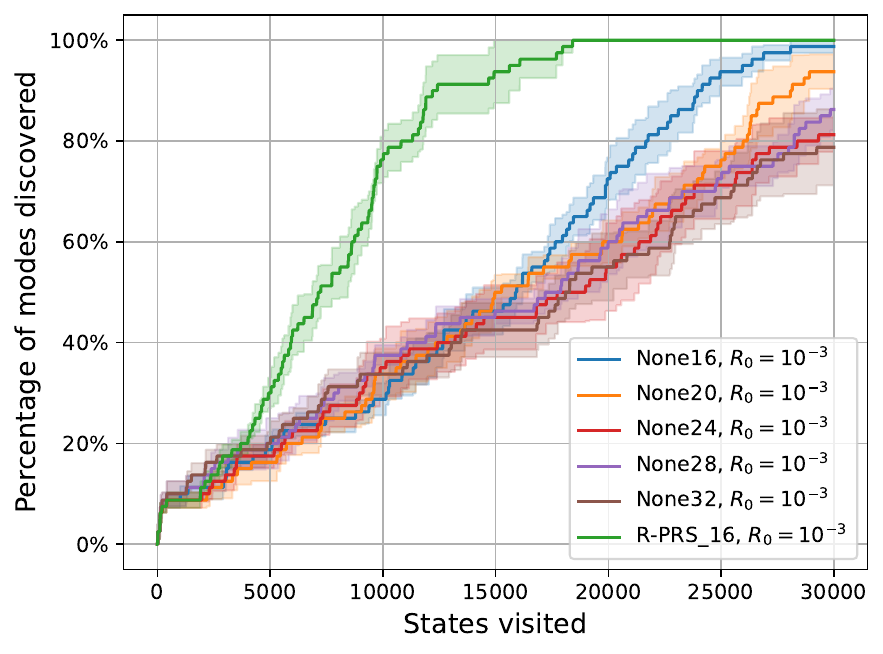}
    \vskip -0.1in
    \caption{Hypergrid domain - States visited vs. percentage of modes discovered during training in a 4-dimensional hypergrid (max = 16 modes) with $H=8$ for different batch sizes, with $R_0=10^{-3}$ (mean and standard error over 5 runs).}
    \label{hypergrid_1e-3_NoneAllBatches}
  \end{minipage}
  % \caption{Side-by-side images}
  % \label{fig:side_by_side_images}
\end{figure*}

We present the results of an experiment with $R_0=10^{-3}$, one of the more difficult settings. 
For each batch, the agent draws an equal number of trajectories from both the online policy and the replay buffer ($16$ trajectories each). Figure~\ref{hypergrid_1e-3_Allmethods} shows the evolution of the number of modes discovered as a function of training samples. R-PRS discovers all the modes relatively quickly compared to the random sampling and no replay buffer settings. Figure~\ref{hypergrid_empl1_1e-3} shows that the R-PRS technique exhibits faster convergence towards the true reward distribution, compared to the other methods. Similar results in a relatively easier setting, $R_0=10^{-2}$, are shown in Appendix~\ref{A1}. This suggests that GFlowNets are more capable of exploration when the learner is repeatedly exposed to more promising samples (samples with high rewards).

To further evaluate the impact of the experience replay sample size, we plot mode discovery as a function of the number of trajectories sampled from the replay buffer for use with R-PRS. The agent samples $16$ trajectories from the online policy. We vary the number of trajectories from older policies from  $4$ to $16$. Figure~\ref{hypergrid_1e-3_RPRS} shows that increasing the number of older trajectories sampled from the replay buffer helps the learner to discover modes more quickly. 
We can observe similar kinds of results in a relatively easier setting, $R_0=10^{-2}$ as shown in Appendix~\ref{A1}.

To analyze whether the improvement in performance is due to the increased sample size from the experience replay buffer, we plot the modes discovered as a function of increasing batch size. When using no replay buffer, we varied the batch size from $16$ to $32$ and included R-PRS for comparison. In Figure~\ref{hypergrid_1e-3_NoneAllBatches}, we observe that solely increasing the batch size (number of samples) negatively affects performance, thereby confirming that drawing high-reward samples from the replay buffer yields better results, compared to simply drawing additional samples from the online policy. Similar results can be observed in a relatively simpler setting, with $R_0=10^{-2}$, as shown in Appendix~\ref{A1}.

As shown in  Appendix~\ref{A4}, similar results can be observed when mode discovery is plotted as a function of both the size of the experience replay and the sample size of the replay buffer.

\subsection{Molecule synthesis}

We carry out further analysis in a large-scale, a molecular synthesis environment, where the objective is to generate small molecules that have low binding affinity to a pre-specified target. In this environment, the reward function is the binding affinity of a candidate molecule to the target protein. 

The objective is to generate a diverse set of molecules that exhibit high reward. The environment has approximately $10^{16}$ states, and the number of available actions ranges from 100 to 2000, depending on the agent's current state. Inspired by the work of \citeauthor{bengio2021flow} and following the framework proposed by \citeauthor{jin2019chapter}, we adopt a method for molecule generation that utilizes a predefined vocabulary of building blocks. The process involves constructing graphs through iterative addition. Each action corresponds to selecting a specific block to attach and determining the attachment point. This construction process gives rise to a directed acyclic graph (DAG), as multiple action sequences can lead to the same resulting graph. The details about the reward signal in this environment are shown in Appendix~\ref{mols_avgreward}.

\begin{figure}[ht]
\vskip 0.2in
\begin{center}
\centerline{\includegraphics[scale=0.4]{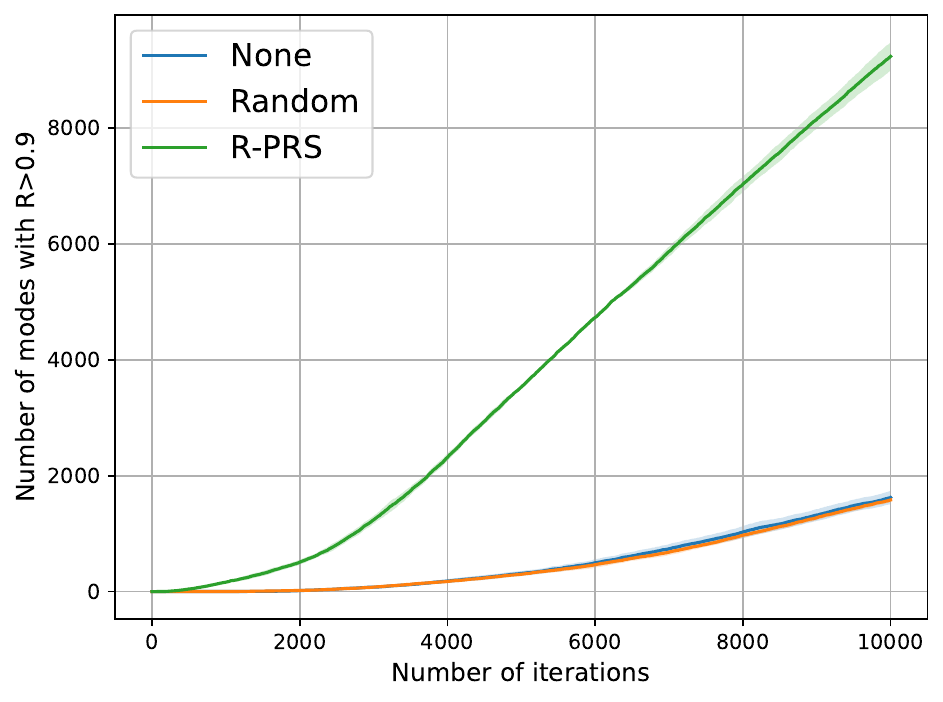}}
\caption{Molecule synthesis environment - Number of iterations vs. the number of modes discovered with a reward at least $r>0.9$ during training for R-PRS, Random sampling from replay buffer, and no replay buffer (mean and standard error over 3 runs).}
\label{mols_allmodes}
\end{center}
\vskip -0.5in
\end{figure}

We tested the impact of the replay buffer on this large-scale environment by experimenting with all three training techniques introduced earlier. Figure~\ref{mols_allmodes} shows the number of modes discovered by each of these techniques. We identify all the candidates with rewards of more than $0.9$ as modes. It is clear that  R-PRS performs significantly better in terms of mode discovery. The average reward during the training is also better for R-PRS, as shown in Appendix~\ref{mols_avgreward}. Concurrent work by \citeauthor{shen2023towards} proposes prioritized replay training with high-reward samples as well. The authors claim that the performance of GFlowNets improved with the inclusion of experience replay.  Figure~\ref{mols_allmodes} shows an interesting insight: the performance of the model without the replay buffer and the performance of the model with random sampling from the replay buffer is almost identical. This further ascertains that increasing the access to promising trajectories during training is what improves the performance, not just the use of the replay buffer.

\section{Discussion}
In this paper, we conducted an empirical study of the effect of using  an experience replay buffer containing past experience in GFlowNets training. Our empirical results show that using a prioritized replay which encourages the use of high-reward trajectories provides a performance boost in terms of mode discoverability as well as training speed. 

This, in turn, led to an increase in the diversity of candidate solutions without compromising on training convergence. We have also shown that increasing the size of the experience replay and of the replay buffer sample during training has a positive impact on the performance.

While our experimentation was limited to a couple of variants of experience replay,  additional variations may further improve learning performance. Investigating other methods for improving learning speed and stability from the RL literature may also bring GFlowNet performance improvements.

\section*{Acknowledgements}
We gratefully acknowledge the generous funding support received for this project. We would like to express our sincere gratitude to the Fonds Recherche Quebec for their FACS-Acquity grant and to the National Research Council of Canada. Their financial contributions have played a vital role in making this research possible.

% \textbf{Do not} include acknowledgements in the initial version of
% the paper submitted for blind review.

% If a paper is accepted, the final camera-ready version can (and
% probably should) include acknowledgements. In this case, please
% place such acknowledgements in an unnumbered section at the
% end of the paper. Typically, this will include thanks to reviewers
% who gave useful comments, to colleagues who contributed to the ideas,
% and to funding agencies and corporate sponsors that provided financial


\bibliography{example_paper}
\bibliographystyle{icml2023}

%%%%%%%%%%%%%%%%%%%%%%%%%%%%%%%%%%%%%%%%%%%%%%%%%%%%%%%%%%%%%%%%%%%%%%%%%%%%%%%
%%%%%%%%%%%%%%%%%%%%%%%%%%%%%%%%%%%%%%%%%%%%%%%%%%%%%%%%%%%%%%%%%%%%%%%%%%%%%%%
% APPENDIX
%%%%%%%%%%%%%%%%%%%%%%%%%%%%%%%%%%%%%%%%%%%%%%%%%%%%%%%%%%%%%%%%%%%%%%%%%%%%%%%
%%%%%%%%%%%%%%%%%%%%%%%%%%%%%%%%%%%%%%%%%%%%%%%%%%%%%%%%%%%%%%%%%%%%%%%%%%%%%%%
\newpage
\appendix
\onecolumn
\section{Appendix} 
We use Python 3.9 \cite{10.5555/1593511} to run all our experiments. We implemented all the ML models using PyTorch \cite{paszke2019pytorch}. Following  \citeauthor{bengio2021flow}, we use the AutoDock Vina library~\cite{trott2010autodock}  for binding energy estimation and the RDKit library~\cite{landrum2006rdkit}  for chemistry routines.
We use NVidia RTX 8000 GPUs with 4 CPU cores in a cluster to train on the molecule synthesis environment and single-core CPUs to train on the Hypergrid environment.

\subsection{Additional Hypergrid Experiments}\label{A1}

\begin{figure*}[h]
  \centering
  \begin{minipage}[t]{0.3\textwidth}
    \centering
    \includegraphics[width=\textwidth]{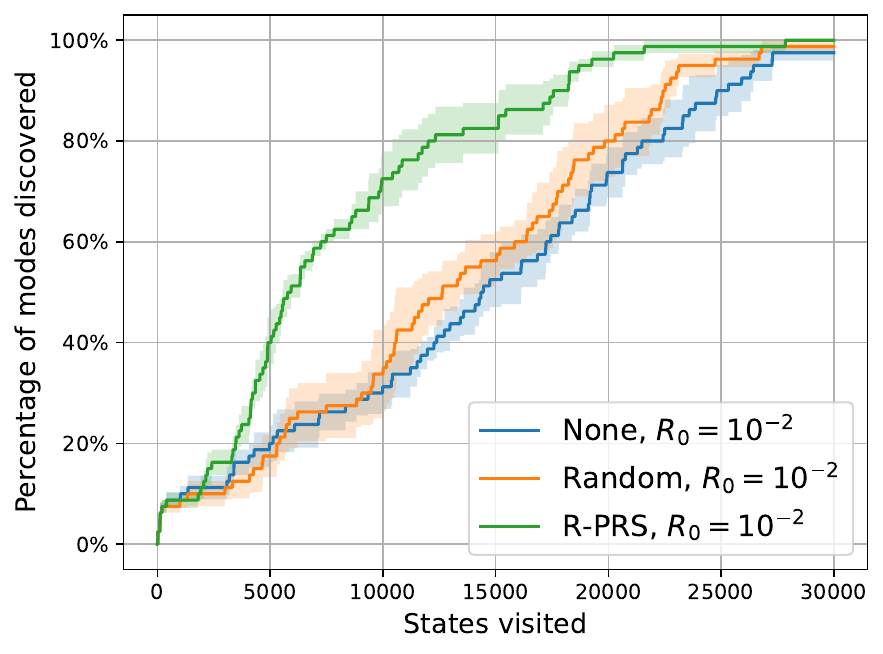}
    \vskip -0.1in
    \caption{Hypergrid domain - States visited vs. percentage of modes discovered during training in a 4-dimensional hypergrid (max = 16 modes) with $H=8$ for all three proposed training techniques, with $R_0=10^{-2}$ (mean and standard error over 5 runs).}
    \label{hypergrid_1e-2_Allmethods}
    % \vspace{-50pt}
  \end{minipage}
  \hfill
  \begin{minipage}[t]{0.3\textwidth}
    \centering
    \includegraphics[width=\textwidth]{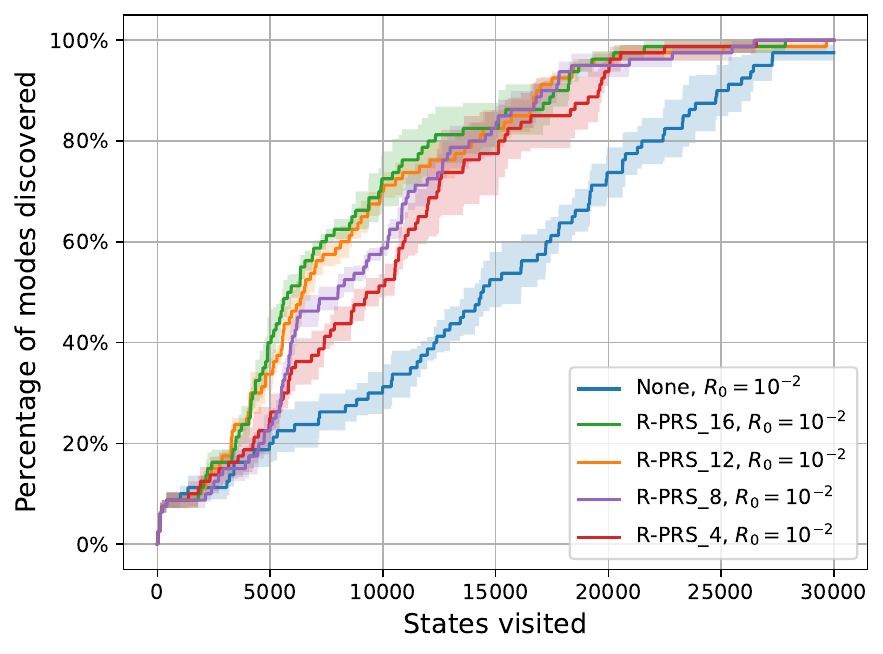}
    \vskip -0.1in
    \caption{Hypergrid domain - States visited vs. percentage of modes discovered during training in a 4-dimensional hypergrid (max = 16 modes) with $H=8$ for varying sample sizes for R-PRS, with $R_0=10^{-2}$ (mean and standard error over 5 runs).}
    \label{hypergrid_1e-2_RPRS}
    % \vspace{-50pt}
  \end{minipage}
  % \vskip -0.2in
  \hfill
  \begin{minipage}[t]{0.3\textwidth}
    \centering
    \includegraphics[width=\textwidth]{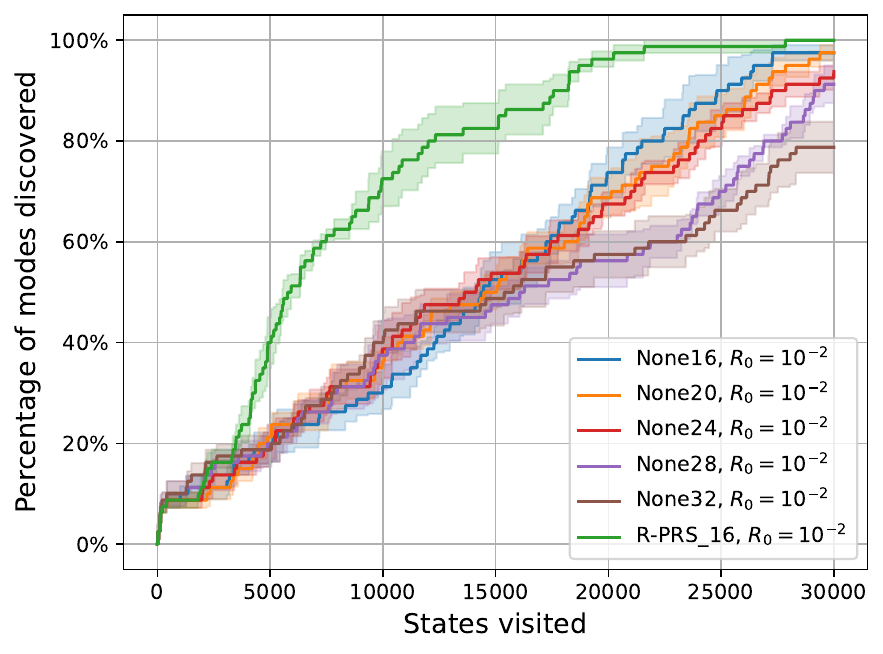}
    \vskip -0.1in
    \caption{Hypergrid domain - States visited vs. percentage of modes discovered during training in a 4-dimensional hypergrid (max = 16 modes) with $H=8$ for different batch sizes of only samples from the online policy (no replay buffer) and R-PRS, with $R_0=10^{-2}$ (mean and standard error over 5 runs).}
    \label{hypergrid_1e-2_NoneAllbatches}
  \end{minipage}
  % \caption{Side-by-side images}
  % \label{fig:side_by_side_images}
\end{figure*}

In the \textit{Experiments} section of the main paper, we presented the influence of different experience replay variants on the GFlowNets training process, in the difficult setting of $R_0=10^{-3}$. We also conducted similar tests on a relatively easier setting, $R_0=10^{-2}$. The results are as shown in Figure~\ref{hypergrid_1e-2_Allmethods}. As before, R-PRS performs better in terms of mode discovery, followed by the replay buffer with random sampling. 
We also plot the number of modes discovered as a function of the replay buffer sample size and of the batch size, in Figure~\ref{hypergrid_1e-2_RPRS} and Figure~\ref{hypergrid_1e-2_NoneAllbatches} respectively. 

\begin{figure}[ht]
% \vskip 0.2in
\begin{center}
\centerline{\includegraphics[scale=0.5]{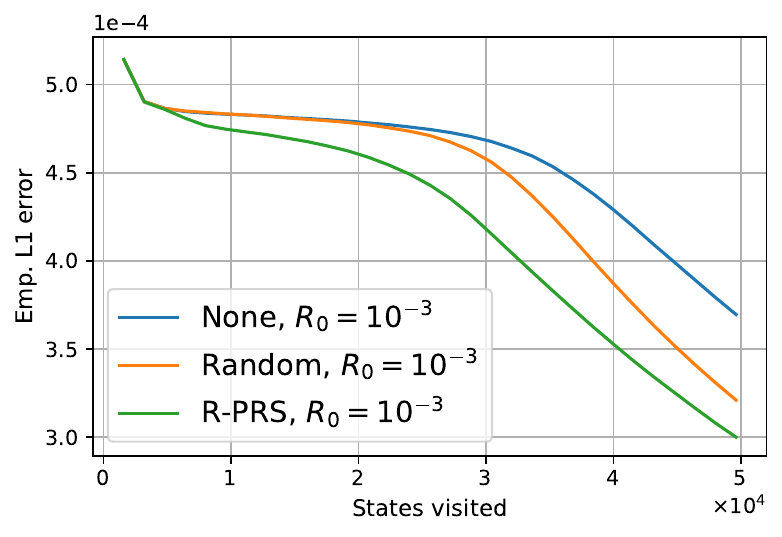}}
\caption{Hypergrid domain - Empirical L1 error vs States visited during training in a 4-dimensional hypergrid with $H=8$ and $R_0=10^{-3}$ (mean over five independent runs).}
\label{hypergrid_empl1_1e-3}
\end{center}
\vskip -0.2in
\end{figure}

Figure~\ref{hypergrid_empl1_1e-3} illustrates the impact of the replay buffer on the training efficiency of GFlowNets. The empirical L1 error, measured between the true reward distribution and the learned reward distribution, is used as the evaluation metric.  R-PRS converges faster than the other techniques towards the true reward distribution.

\subsection{Mode discovery as a function of  replay buffer size and batch size}\label{A4}

\begin{figure*}[h]
  \centering
  \begin{minipage}[t]{0.3\textwidth}
    \centering
    \includegraphics[width=\textwidth]{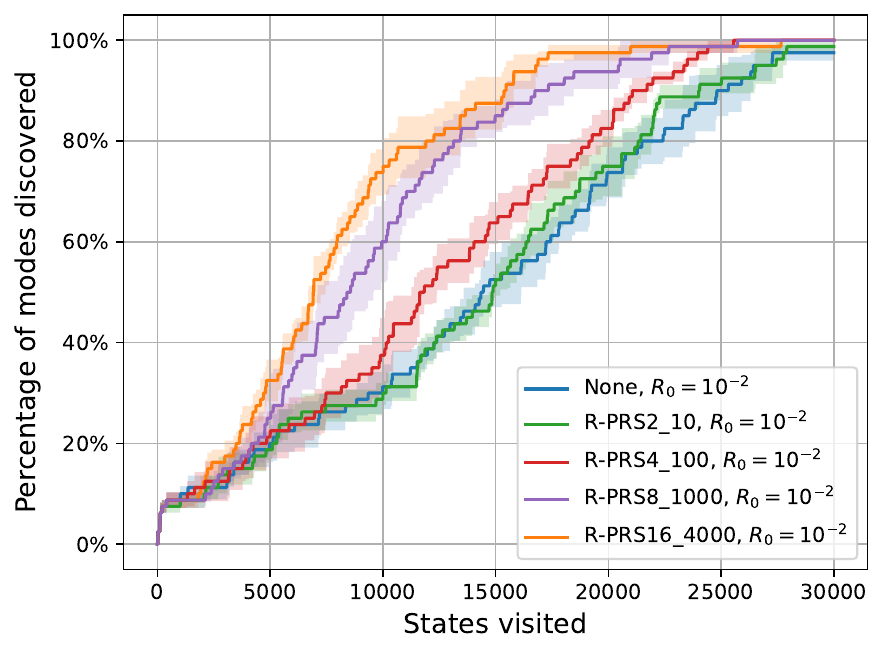}
    \vskip -0.1in
    \caption{Hypergrid domain - States visited vs. percentage of modes discovered during training in a 4-dimensional hypergrid (max = 16 modes) with $H=8$ as a function of increasing the experience replay sample size and the replay buffer size for R-PRS replay technique, with $R_0=10^{-2}$ (mean and standard error over 5 runs).}
    \label{hypergrid_1e-2_RPRS_ndim4_varRSS}
    % \vspace{-50pt}
  \end{minipage}
  \hfill
  \begin{minipage}[t]{0.3\textwidth}
    \centering
    \includegraphics[width=\textwidth]{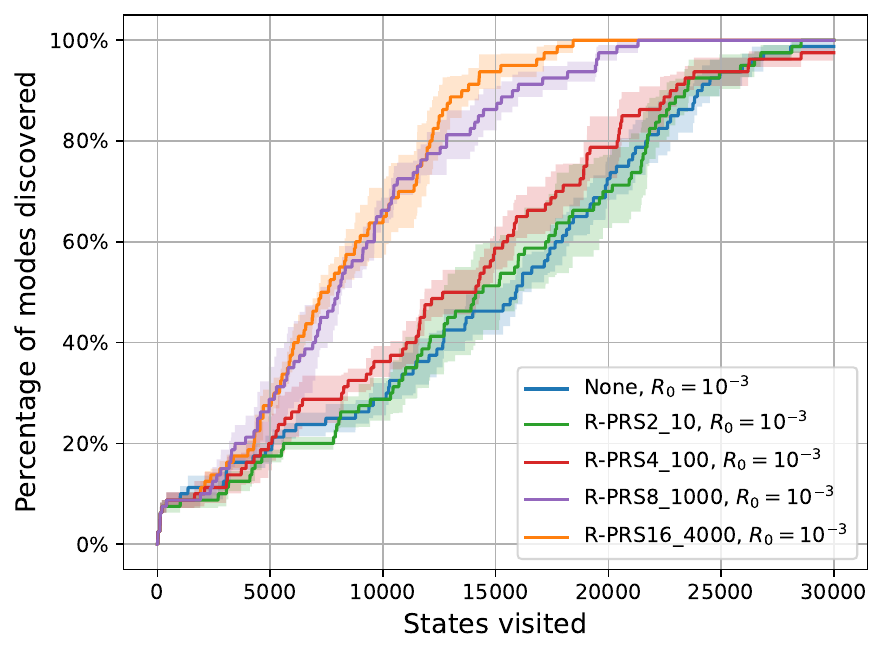}
    \vskip -0.1in
    \caption{Hypergrid domain - States visited vs. percentage of modes discovered during training in a 4-dimensional hypergrid (max = 16 modes) with $H=8$ as a function of increasing the experience replay sample size and the replay buffer size for R-PRS replay technique, with $R_0=10^{-3}$ (mean and standard error over 5 runs).}
    \label{hypergrid_1e-3_RPRS_ndim4_varRSS}
    % \vspace{-50pt}
  \end{minipage}
  % \vskip -0.2in
  \hfill
  \begin{minipage}[t]{0.3\textwidth}
    \centering
    \includegraphics[width=\textwidth]{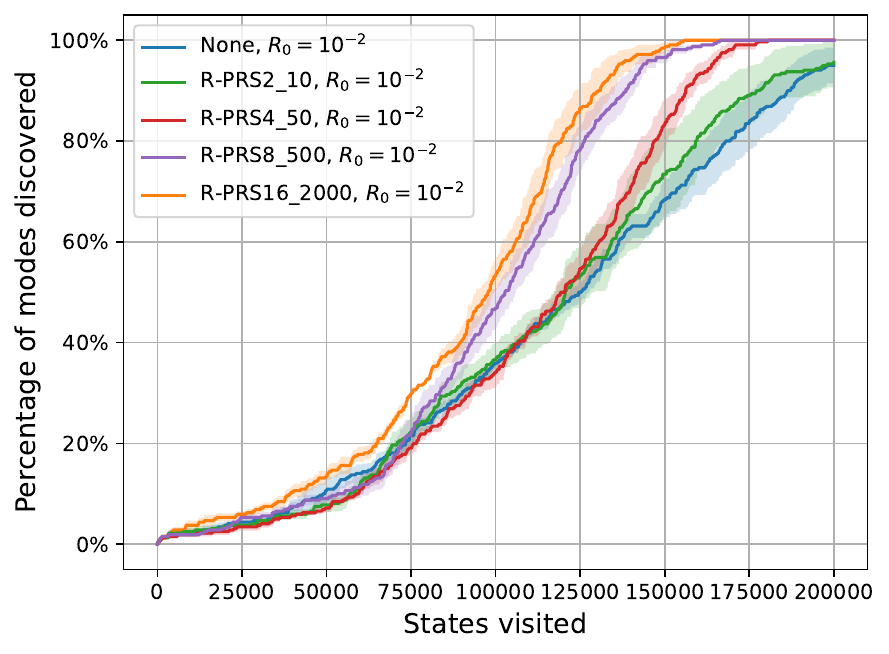}
    \vskip -0.1in
    \caption{Hypergrid domain - States visited vs. percentage of modes discovered during training in a 6-dimensional hypergrid (max = 64 modes) with $H=8$ as a function of increasing replay sample size and experience replay size for R-PRS replay technique, with $R_0=10^{-2}$ (mean and standard error over 5 runs).}
    \label{hypergrid_1e-2_RPRS_ndim6_varRSS}
  \end{minipage}
\end{figure*}

So far, we presented the mode discovery as a function of increasing replay buffer sample size. In this section, we show the mode discovery as a function of both increasing replay buffer sample size and increasing experience replay size. Figure~\ref{hypergrid_1e-2_RPRS_ndim4_varRSS} and Figure~\ref{hypergrid_1e-3_RPRS_ndim4_varRSS} shows the outcome of increasing both the entities in easy and difficult setting respectively. We vary our experiment with the replay buffer sample size from $2$ to $16$ samples per batch and the experience replay size from $10$ to $4000$ samples in a $4$ dimensional hypergrid. 

We conduct a similar experiment in a $6$ dimensional hypergrid and the results are as shown in Figure~\ref{hypergrid_1e-2_RPRS_ndim6_varRSS}. From the results obtained in both 4D and 6D settings, it can be observed that as the buffer size and sample size increase, there is an acceleration in mode discovery.

\subsection{Additional Molecule Synthesis} \label{mols_avgreward}
\begin{figure}[ht]
\vskip 0.2in
\begin{center}
% \centerline{\includegraphics[width=\columnwidth]{Plots/1e-3_AllModes.pdf}}
\centerline{\includegraphics[scale=0.5]{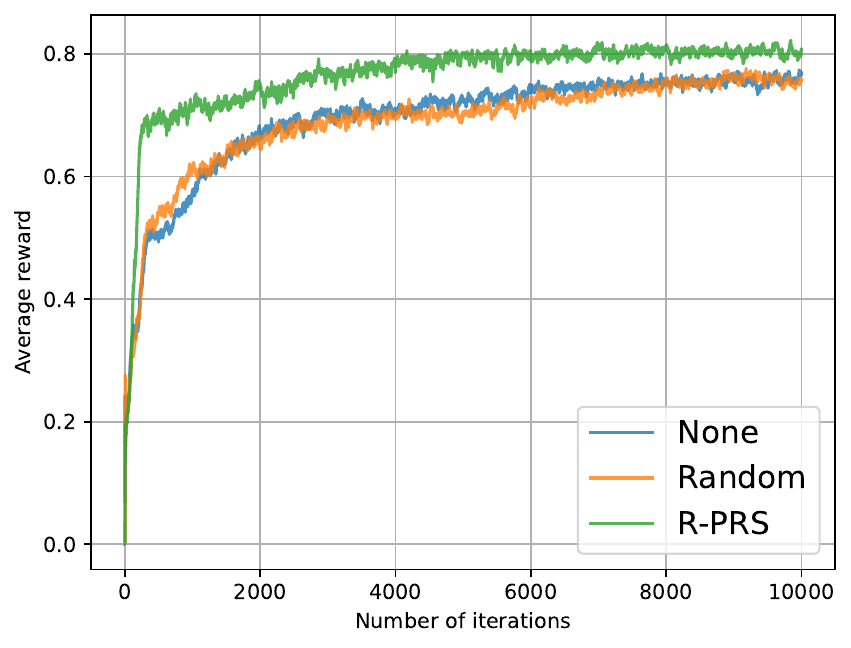}}
\caption{Molecule synthesis environment - Number of iterations vs. average reward during training for different training techniques namely R-PRS, Random sampling from replay buffer, and no replay buffer (mean and standard error over 3 runs).}
\label{mols_avgrew}
\end{center}
\vskip -0.2in
\end{figure}
The reward signal in this environment is determined by the binding energy required to attach a molecule to a target protein. However, computing binding energies is computationally expensive. To address this challenge, \citeauthor{bengio2021flow} developed a pretrained proxy model that predicts the binding energy for a given molecule and target protein. The proxy employs a message-passing neural network (MPNN) \cite{gilmer2017neural} parameterized over the atom graph. To train the proxy model, a semi-curated semi-random dataset of 300k molecules is utilized.

We use the pretrained proxy developed by \citeauthor{bengio2021flow} for binding energy estimation. We use the subtrajectory balance objective introduced by \citeauthor{madan2022learning} for our GFlowNet training in the molecule environment. We use the training architecture, hyperparameters, and dataset as provided by \citeauthor{bengio2021flow} for our experiments. The binding energy scores, i.e., the reward for the candidates are computed with AutoDock \cite{trott2010autodock}. All the experiments are run on 3 independent seeds and the mean and standard error are reported in the plots. To test the hyperparameter robustness, we tried various batch sizes $64$, $128$, and $256$, varied the replay buffer sampling size from $64$ to $128$, and tested different replay buffer sizes, $100$, $1000$, and $5000$ samples. We trained all the models for $10000$ iterations.

As shown in the \textit{Experiment} section of the main paper, we conducted experiments with all three training techniques proposed. Figure~\ref{mols_avgrew} displays the average reward for each technique throughout the number of iterations. It is evident that R-PRS outperforms the other two techniques.

%%%%%%%%%%%%%%%%%%%%%%%%%%%%%%%%%%%%%%%%%%%%%%%%%%%%%%%%%%%%%%%%%%%%%%%%%%%%%%%
%%%%%%%%%%%%%%%%%%%%%%%%%%%%%%%%%%%%%%%%%%%%%%%%%%%%%%%%%%%%%%%%%%%%%%%%%%%%%%%

\end{document}